\documentclass[10pt,twocolumn]{article}

\usepackage[letterpaper,top=0.70in,bottom=0.76in,left=0.67in,right=0.67in,headsep=0.16in]{geometry}
\usepackage[T1]{fontenc}
\usepackage{newtxtext,newtxmath}
\usepackage{microtype}
\usepackage{amsmath}
\usepackage{graphicx}
\usepackage{booktabs,tabularx,array}
\usepackage{enumitem}
\usepackage{natbib}
\usepackage{xurl}
\usepackage[dvipsnames]{xcolor}
\usepackage[font=small,labelfont=bf]{caption}
\usepackage{dblfloatfix}
\usepackage{placeins}
\usepackage{flushend}
\usepackage{titlesec}
\usepackage{fancyhdr}
\usepackage[hidelinks,unicode]{hyperref}

\definecolor{linkblue}{HTML}{285C7A}
\hypersetup{
  colorlinks=true,
  linkcolor=linkblue,
  citecolor=linkblue,
  urlcolor=linkblue,
  pdftitle={A Calibrated Test of Internal Action Maps: State Signals Without Global Affine Closure},
  pdfauthor={Dekun Yang},
  pdfsubject={Mechanistic interpretability and compositional state transitions},
  pdfkeywords={mechanistic interpretability, causal intervention, compositional generalization}
}


\titleformat{\section}{\large\bfseries}{\thesection.}{0.45em}{}
\titleformat{\subsection}{\normalsize\bfseries}{\thesubsection}{0.45em}{}
\titlespacing*{\section}{0pt}{1.35ex plus .25ex minus .15ex}{0.65ex}
\titlespacing*{\subsection}{0pt}{1.05ex plus .2ex minus .1ex}{0.45ex}
\setlist{nosep,leftmargin=*}
\begin{document}

\twocolumn[{
\begin{center}
  \vspace*{-1.3em}
  {\LARGE\bfseries A Calibrated Test of Internal Action Maps:\par}
  \vspace{0.15em}
  {\LARGE\bfseries State Signals Without Global Affine Closure\par}
  \vspace{0.65em}
  {\normalsize Dekun Yang\textsuperscript{1,*}\par}
  {\small \textsuperscript{1}Zhejiang University\par}
  {\footnotesize *Correspondence: \nolinkurl{pauliyangwork@gmail.com}\par}
  {\footnotesize ORCID: Dekun Yang: \href{https://orcid.org/0009-0002-3496-3596}{0009-0002-3496-3596}\par}
  \vspace{0.35em}
  {\footnotesize Preprint; evidence cutoff 12 August 2026\par}
  \vspace{0.65em}
  \begin{minipage}{0.95\textwidth}
    \small
    \textbf{Abstract.} A hidden state signal can be decodable or causally usable without supporting a reusable action map. We test whether action maps fitted without a source reach its natural post-action activation and compose. We organize the tests as an evidence lattice and validate the geometric branch on a known affine \(S_5\) carrier: all held-source folds pass one-step, composition, inverse, decoding, and commutativity gates. Structured curvature and held-domain conjugacy raise error monotonically, but only 23/30 strongest cells flip a closure gate, bounding rather than universalizing calibration. In post-trained \texttt{Qwen/Qwen3-4B}, frozen final-token h28 affine maps have mean held-entity error \(.519\), versus \(.398\) for within-test-domain cross-fit. Seven randomized entity splits and map geometry do not support a purely entity-specific account. Earlier h4/h16 layers fit one-step transitions better, but h4 conflict-state decoding is weak and lexical controls remain unresolved. Three matched intervention datasets regenerated from one frozen checkpoint show causal effects only at h28/h36. Outcome-aware refitting improves h28 one-step error to \(.474\) (\(.469\) with weighting), yet no refit passes composition. Learned finite worlds likewise preserve relative algebraic signals or shared charts without held-source affine closure. Within the tested carriers, state availability, causal use, local geometry, and reusable closure are separable. The result is limited to one pretrained model, sampled final-token layers, two finite worlds, and the tested affine or diagnostic function classes.

    \vspace{0.45em}
  \end{minipage}
\end{center}
\vspace{0.8em}
}]

\section{Introduction}\label{introduction}

Models that track a changing world must bind entities to their current attributes, update those attributes after actions, and carry the consequences forward. Language-model activations can encode entity states, and interventions on those activations can change later predictions \citep{li2021implicit, kim2023entity}. Other work has identified computations that route or retrieve state information during fine-tuning and belief tracking \citep{prakash2024finetuning, prakash2026lookbacks}. Together, these findings establish the presence and relevance of a state signal. They stop short of showing that each action implements a stable function that transfers to held-out sources and composes with other actions.

Several weaker observations can look like evidence for such an operator. A probe may decode an off-manifold activation, or an affine map may beat a mean direction while still missing much of the natural target displacement. Correct and reversed action orders can also be ranked when neither endpoint is accurate. Static relation maps, task vectors, and function vectors make the operator hypothesis plausible \citep{hernandez2024linearity, hendel2023task, todd2024function}. None of these observations alone establishes closure over changing world states.

Our measurement framework is therefore a lattice rather than a sequential ladder. Once the availability of state information is established, causal local use and reusable closure/algebraic laws can be tested independently. Geometry without a detected output effect remains informative, as does causal use without held-source closure. Neither is the full claim. Under a behaviorally eligible task, only evidence from both branches licenses the phrase ``causally faithful reusable operator.'' This partial order avoids implying that causal patching must precede every geometric diagnostic.

We study one controlled language-model setting alongside two exact finite worlds. The grounded experiments freeze the post-trained thinking/non-thinking \texttt{Qwen/Qwen3-4B} checkpoint for an Alchemy state-update task. The other experiments train decoder-only Transformers in a 121-state \(\mathbb{Z}_{11}^2\) system and the 120-state permutation group \(S_5\). Alchemy retains language-model context dependence and permits paired interventions. The finite worlds provide complete transition tables, commuting labels, inverses, and training trajectories.

A failed gate immediately raises a construct-validity question: would the test pass when the operator is known to exist? We answer it with an end-to-end positive control, a known contextual \(S_5\) carrier evaluated under the same held-source logic. A parity-support split, smooth observation curvature, and held-domain coordinate conjugacy then stress that carrier. Exact recovery checks the implementation; the structured stresses probe behavior under misspecification. Their calibration is informative but limited, so we do not treat success on one favorable noise family as evidence of universal power.

The grounded results separate reconstruction, causal use, and carrier identity. Within-test-domain cross-fitting improves h28 reconstruction, but randomized entity splits do not support maps dominated by entity identity. Action identity instead dominates the tested parameter geometry. One-step reconstruction is best at h4/h16, whereas independently regenerated paired interventions are effective at h28/h36. The lexical and state-probe controls do not settle whether h4 is a surface carrier. Finally, a frozen bridge associates most of the apparent h28 one-step improvement with final refitting rather than inverse-displacement weighting or a larger affine function class. Those refit maps still fail composition.

Methodologically, the study provides a calibrated evidence lattice built around natural endpoints, held sources, distinct labels for negative and blocked outcomes, and independent audits of every formal follow-up. Empirically, the tested representations separate state availability, causal use, action-dominant parameter geometry, one-step specification sensitivity, relative algebraic discrimination, and global affine closure. These findings apply to the tested carriers and hypotheses. They are not a general denial of internal operators.

\section{Related Work}\label{related-work}

\subsection{State tracking and causal state use}\label{state-tracking-and-causal-state-use}

Entity-tracking studies show that hidden activations can contain dynamic world information. \citet{li2021implicit} combined decoding with intervention, while \citet{kim2023entity} studied behavioral tracking across entities and sequence conditions. In procedural text, Neural Process Networks update a designed entity-memory state with learned action operators \citep{bosselut2018process}. Later work examined how fine-tuning reuses entity-tracking mechanisms and how lookback operations bind beliefs to earlier observations \citep{prakash2024finetuning, prakash2026lookbacks}. Together, these studies address availability, binding, causal relevance, and architecturally specified state transformation. Our closure branch asks something else: whether an action-conditioned map recovered from an existing model's residual stream predicts the natural post-action representation of sources excluded from fitting.

\subsection{Linear transformations and composed functions}\label{linear-transformations-and-composed-functions}

Affine relation decoders can map subjects toward objects and support causal edits \citep{hernandez2024linearity}. Task and function vectors can summarize computations induced by demonstrations and influence output behavior \citep{hendel2023task, todd2024function}. Work on continuous latents and compositional primitives extends this idea to multi-step computation \citep{hao2025coconut, lippl2026primitives}.

\citet{khandelwal2026compose} study two-hop factual functions \(g(f(x))\), identify residual-stream signatures of the intermediate variable, and compare direct with compositional mechanisms. Our objects, representations, closure test, and causal claims differ. We condition maps on explicit world actions, score natural hidden endpoints, hold out entities or source states from fitting, and interpret relative laws only after absolute reconstruction. An intermediate representation can support function evaluation without forming a global affine action algebra.

\subsection{Algebraic structure and world models}\label{algebraic-structure-and-world-models}

Under suitable tasks and interfaces, structural studies show that algebraic computation can be learned or imposed. \citet{li2025state} identify associative and parity-associative mechanisms in Transformers trained on permutation words. \citet{an2026homomorphism} connect representation-level homomorphism error with compositional generalization. \citet{lee2026falsifier} evaluates a structured recurrent interface using held-out transition pairs, drift, homomorphism, and commutator diagnostics. These constructions motivate our positive carrier and show why failure in an ordinary residual stream is not an impossibility result.

Behavior supplies an orthogonal standard. Transformers trained on Markov decision processes can encode transition dynamics \citep{chen2024transition}, yet strong local predictions may coexist with an incoherent recovered world model \citep{vafa2024world}. Probe controls raise a related concern: a measurement can reward its own fitting capacity instead of the intended construct \citep{hewitt2019probes}. A dated search log records databases, query families, screening bounds, and a closest-work feature matrix. Within that documented search through 12 August 2026, we found no prior study combining known-algebra calibration, held-source natural endpoints, matched interventions, law tests, behavioral eligibility, and independent artifact audits across an existing language model and exact finite systems. This is a search-bounded description of the integrated protocol, not a topic-level priority claim.

\section{Evidence Framework}\label{evidence-framework}

\subsection{From a state signal to an action map}\label{from-a-state-signal-to-an-action-map}

Let \(h(s,c)\) be the representation of world state \(s\) in context or history \(c\). For action \(a\), the primary family is

\[
A_a(h)=W_a h+b_a.
\]

Interpolating the observed pairs is not enough. A reusable \(W_a\) must transfer across entities, contexts, histories, or source states withheld from fitting. For Alchemy, row-relative endpoint error is \(\|A_a(h)-h'\|_2/\|h'-h\|_2\); identity therefore has error one whenever the displacement is nonzero. In the finite worlds, normalized root-mean-square error (NRMSE) divides pooled squared error by a frozen target-variance reference. Both metrics compare the prediction with the natural post-action activation, rather than only checking a decoded label.

Consider a held entity initially in state \texttt{green}, followed by \texttt{fill\_red} and \texttt{fill\_blue}. A state probe checks whether the final activation decodes as blue. An order test compares \(A_{\mathrm{blue}}A_{\mathrm{red}}h\) with the reversed order. Absolute closure instead asks whether the composed point matches the held source's natural final activation. The probe and order test can both succeed while that endpoint remains inaccurate.

State availability is measured separately from causal local use. On held data, a fixed probe is evaluated against a deterministic random-label control. The preregistered residual replacement must change the target-versus-source logit margin more than a matched wrong-state or translation control. Because patching may move the activation off manifold, a successful intervention licenses a causally usable direction, not a naturally traversed transition.

\subsection{Absolute closure and relative laws}\label{absolute-closure-and-relative-laws}

H1 compares one-step reconstruction with identity, mean translation, and random-map controls. The grounded diagnostics also include a residual MLP, radial-basis-function kernel ridge, within-test-domain cross-fit, source-state-label-gated affine bank, and an optimistic in-sample capacity ceiling. Each reference answers a different question, and several are not deployable mechanisms.

H2 scores \(A_b(A_a(h))\) against the natural two-step endpoint and compares it with a directly fitted two-action map. Because the second setter overwrites the first, the setter domain also requires a last-action-only baseline. The frozen grounded decision combines endpoint error, direct-map discrepancy, order effect, and state-probe gain. An order contrast cannot pass H2 on its own.

H3 concerns commutativity. In Alchemy, its inferential units are matched same-entity and disjoint-entity pairs. In the finite worlds, the transition table labels every unordered action pair; AUROC then measures whether normalized commutators rank noncommuting pairs above commuting pairs. H4 applies the ground-truth inverse and scores the return to the source representation. At the checkpoint level, H5 tested whether algebraic violation covaried with behavioral incoherence.

\subsection{Gates, branches, and evidence labels}\label{gates-branches-and-evidence-labels}

\begin{table*}[t]
  \centering
  \caption{\textbf{Evidence lattice.} Availability enables two independent
  branches. Only their conjunction under behavioral eligibility licenses the
  strongest operator claim.}
  \label{tab:lattice}
  \footnotesize
  \setlength{\tabcolsep}{4pt}
  \renewcommand{\arraystretch}{1.14}
  \begin{tabularx}{\textwidth}{@{}>{\raggedright\arraybackslash}p{0.16\textwidth}>{\raggedright\arraybackslash}p{0.24\textwidth}>{\raggedright\arraybackslash}p{0.29\textwidth}X@{}}
    \toprule
    Node & Measurement & Licensed statement & Does not establish \\
    \midrule
    State availability & Held-data decoding versus label controls
      & State information is available to the decoder
      & Causal use or a natural update path \\
    Causal-use branch & Matched intervention versus wrong-state/translation controls
      & A state direction is locally usable by the output computation
      & Held-source closure or an on-manifold transition \\
    Closure/law branch & Natural endpoints, then composition/inverse/commutativity
      & The tested map transfers and satisfies the reported law
      & Local output use or an untested carrier \\
    Behavioral eligibility & Task success and state differentiation
      & Model-level interpretation is meaningful
      & A unique internal implementation \\
    Branch conjunction & Causal use plus reusable closure/laws under eligibility
      & A causally faithful reusable operator in the tested carrier
      & Universality across models, layers, tokens, or function classes \\
    \bottomrule
  \end{tabularx}
\end{table*}

All thresholds, partitions, and inferential units were frozen before their corresponding formal results. \texttt{NOT\ SUPPORTED} means that a testable joint gate failed. \texttt{NOT\ TESTED} records an upstream-blocked experiment. \texttt{UNTESTABLE} records a missing logical antecedent or usable observation. \texttt{RIGHT-CENSORED} records an event not observed within a fixed budget. Diagnostic branch results cannot rewrite older frozen verdicts.

\begin{figure*}[t]
  \centering
  \includegraphics[width=\textwidth]{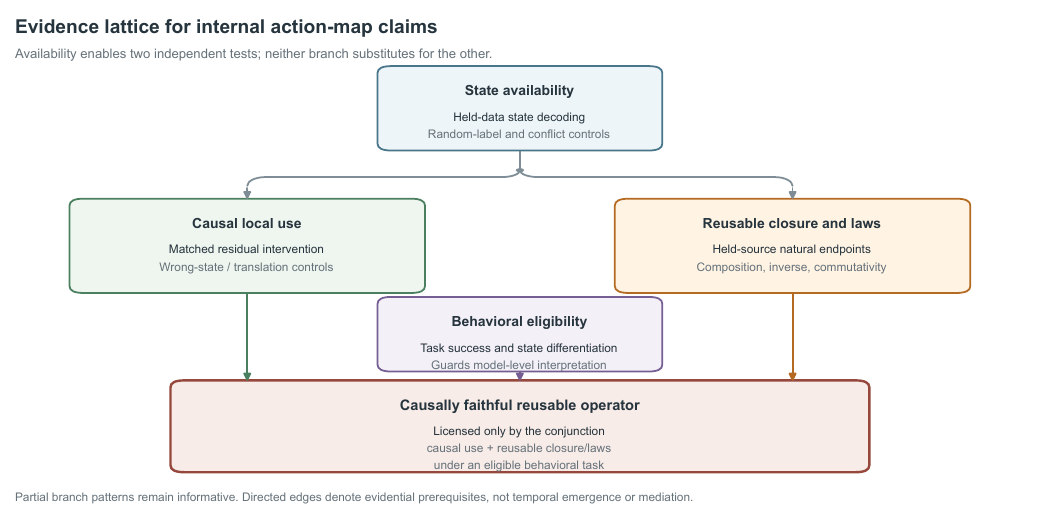}
  \caption{\textbf{Evidence lattice for internal action-map claims.}
  State availability enables independent tests of causal local use and reusable
  closure/laws. Behavioral eligibility guards model-level interpretation.
  Geometry without detected causal use and causal use without closure remain
  distinct partial outcomes; only their conjunction licenses a causally faithful
  reusable operator. Directed edges denote evidential prerequisites, not temporal
  emergence or mediation. The diagram contains no sampled observations.}
  \label{fig:evidence-lattice}
\end{figure*}

\section{Experimental Settings}\label{experimental-settings}

\subsection{Controlled Alchemy and split support}\label{controlled-alchemy-and-split-support}

The grounded experiments used a fixed local snapshot of \texttt{Qwen/Qwen3-4B}, the post-trained thinking/non-thinking checkpoint rather than \texttt{Qwen3-4B-Base}. It has 36 Transformer layers and hidden width 2,560; weights were never updated. Prompts described containers in one of five states: \texttt{empty}, \texttt{red}, \texttt{blue}, \texttt{green}, or \texttt{yellow}. Actions emptied a container or filled it with one color. Representations were taken at the final prompt token.

The original Phase 1 scan froze h28 because it was the earliest sampled location that passed both direct state decoding and a paired state-specific intervention. The split withholds entity identities and generated context/activation instances, not state categories or the two primary templates.

\begin{table*}[t]
  \centering
  \caption{\textbf{Grounded split support.} The protocol holds out entities and
  generated context/activation instances, not state categories or the two primary
  templates.}
  \label{tab:split-support}
  \footnotesize
  \setlength{\tabcolsep}{5pt}
  \renewcommand{\arraystretch}{1.12}
  \begin{tabularx}{\textwidth}{@{}lrrrrX@{}}
    \toprule
    Split & Rows/seed & Entities & State categories & Templates & Role \\
    \midrule
    Train & 2,500 & A--D & all five & 0, 1 & Map fitting \\
    Validation & 500 & E & all five & 0, 1 & Rank/ridge selection \\
    Test & 1,000 & F--G & all five & 0, 1 & Held-entity/context evaluation \\
    \bottomrule
  \end{tabularx}
\end{table*}

For each of three dataset-seed regenerations from the same frozen checkpoint, H1 fits five action maps from 500 train, 100 validation, and 200 test pairs per action. H2 covers all 20 ordered distinct-action sequences and collapses reverse directions into ten semantic pairs for inference. H3 uses matched same-entity and disjoint-entity pairs. The planned natural-language inverse test was stopped when the behavior manipulation failed.

\subsection{Known-algebra calibration and sampled-layer follow-up}\label{known-algebra-calibration-and-sampled-layer-follow-up}

The positive control represents each of the 120 states of \(S_5\) by its \(5\times5\) permutation matrix plus a seven-dimensional repeat-specific context carrier that actions leave unchanged. A seeded orthogonal embedding maps the carrier into 64 observed dimensions. Ten transpositions act linearly on the permutation component and identically on context. Three outer folds use 80 source states and repeats 0--7 for chart/map fitting; 40 states and repeats 8--11 remain excluded. The \(S_5\) permutation matrices span \(1^2+4^2=17\) dimensions (trivial plus standard representation); adding seven context dimensions gives 24, and removing the constant direction absorbed by the affine bias yields the training-determined chart rank 23. A second embedding seed provides an independent reproduction.

Phase 6 added target-side Gaussian nuisance at signal-RMS ratios \(0,.05,.10,.20,.40,.80\), 20 nuisance seeds per ratio, without moving thresholds. It also froze h28 references and scanned h4, h16, h28, and h36 with full and rank-1/4/16/64 residual affine maps. Phase 7 then evaluated h4/h16 on all ordered action pairs with the unchanged composition construct; single-map choices were imported without sequence-data selection.

\subsection{Structured diagnostics and attribution bridge}\label{reviewer-v2-diagnostics-and-attribution-bridge}

Phase 8 was prospectively frozen against the Phase 6/7 artifacts. Structured calibration rebuilt the carrier under five embeddings. It tested an even-to-odd permutation support split, a common smooth quadratic/tanh observation warp, and a held-domain orthogonal conjugacy at strengths \(0,.05,.10,.20,.40,.80\). The latter two preserve an exact latent action while progressively misspecifying a single observed-space affine map.

Entity/context conditioning used h28 only. Stable-hash sampling retained 80 rows per entity-action cell. Seven cyclic outer splits fit four entities, selected on a fifth, and held out two; a five-fold within-entity reference used the same row budget. Residual affine parameter distance combined Frobenius matrix and intercept distance. Because entity, token, and context features covary, even a positive result would not identify a pure entity mechanism.

Lexical controls fitted true and permuted-label state probes at h4/h16/h28/h36, including a conflict stratum in which the last mentioned state word disagreed with the queried current state. A token-only carrier averaged the last 16 static input embeddings. Neutral clauses inserted requested distances of 0, 16, or 64 tokens before the query, after which the same layer-local map protocol was applied.

Metric-aligned fitting weighted training row \(i\) by \(1/\max(\|h'_i-h_i\|_2^2,10^{-8})\), normalized within action and split, and reported train-defined displacement quintiles. Layer controls recorded residual norm, displacement norm, covariance participation ratio \((\mathrm{tr}\,C)^2/\mathrm{tr}(C^2)\), and one-step results after train-fitted scalar RMS normalization. Causal dataset-seed replication regenerated exactly 160 matched intervention pairs for each of three seeds and four sampled layers from the same frozen checkpoint.

After the audited Phase 8 result, an explicitly outcome-aware Phase 8b bridge separated function class, final fitting rows, and weighting. It compared the original full-or-low-rank train fit, full unweighted train fit, full weighted train fit, and the corresponding full train-plus-validation refits. No test row entered selection or fitting. The unweighted and weighted refit maps were then passed unchanged to the frozen h28 H2 datasets, direct maps, probes, resampling procedures, and gates. Phase 8b is descriptive attribution, not a new confirmation or a revision of frozen H1/H2.

\subsection{Exact learned transition systems}\label{exact-learned-transition-systems}

The first learned world has 121 states \((x,y)\in\mathbb{Z}_{11}^2\) and ten bijections: four translations, coordinate swap, joint negation, and four shears. Of 45 unordered action pairs, 21 commute. Models receive a start state and zero to six actions and predict only the endpoint. Three pre-norm decoder-only Transformer scales, 3.2M, 25.3M, and 85.2M parameters, each use three seeds and \(2^{24}=16.78\) million training examples. Evaluation adds lengths seven to twelve, inverse loops, route-equivalent histories, and all ordered distinct pairs.

The second world uses the 120 states of \(S_5\) and ten self-inverse transpositions; 15 of 45 unordered pairs commute. Medium and large trajectories extend to \(2^{25}=33.55\) million examples and 18 fixed checkpoints. E denotes task/state eligibility, D endpoint-order discrimination, J a shared rigid chart across zero-, one-, and two-action conditions, and K held-source affine closure. Only six medium/large trajectories enter the preregistered cross-world denominator; small models remain capacity controls.

\subsection{Statistics, reproducibility, and auditing}\label{statistics-reproducibility-and-auditing}

The grounded descriptive replicate is the regenerated dataset seed (\(n=3\)) from one frozen checkpoint. Prompts, actions, entities, layers, folds, bootstrap draws, and diagnostic cells are not independent model replicates. Pairwise permutation and bootstrap tests use ten semantic action pairs. Intervention intervals resample 160 matched pairs within each dataset seed; the cross-seed direction is reported separately. Positive-control recovery requires every held-source fold. Synthetic trajectory intervals resample complete training trajectories.

The five preregistered grounded action-level comparisons use Benjamini--Hochberg false-discovery-rate control at \(q=.05\). Follow-up diagnostic families carry their prospectively frozen labels but do not alter Phase 1--7 decisions. Phase 8b is explicitly outcome-aware. Exact package versions are recorded by formal run.

Every formal run verifies frozen input hashes, records configuration, code commit, environment and resource telemetry, and writes an atomic artifact manifest before \texttt{SUCCESS}. Independent auditors do not import the fitting, scoring, bootstrap, or verdict helpers under test. Phase 8 independently rechecked structured masks and gates, entity splits and map distances, lexical probes and carriers, weighted fits, layer statistics, all 1,920 intervention rows, and frozen parity. Phase 8b independently refitted all 75 one-step maps, reproduced all three original H2 summaries, and rebuilt every composition gate. Both remote and locally copied artifacts passed their independent audits.

\section{Results}\label{results}

\subsection{The lattice recovers a known algebra but structured calibration is limited}\label{the-lattice-recovers-a-known-algebra-but-structured-calibration-is-limited}

The exact contextual carrier passed every Phase 6 positive-control gate across three formal and three reproduction folds. Held-state, one-step, two-step, and inverse decoding were 1.000. One-step, two-step, and inverse-cycle NRMSEs were approximately \(3.31\times10^{-8}\), \(5.02\times10^{-8}\), and \(6.55\times10^{-8}\); commuting-pair AUROC was 1.000. The independent embedding retained every gate direction, with all continuous comparisons inside the frozen 5\% reproduction tolerance.

Target-side nuisance produced a smooth specificity curve. At nuisance-to-signal ratios \(.05,.10,.20,.40,.80\), mean one-step NRMSE rose to \(.0096,.0191,.0382,.0764,.1528\); two-step error rose to \(.0136,.0272,.0544,.1090,.2188\). This establishes exact-solution recovery and a graded response to isotropic target perturbation, not power against every structured failure.

Phase 8 supplies the harder distinction. The even-to-odd support split and every zero-strength cell passed. Curved and held-domain families each had Spearman \(\rho=1.000\) between strength and median one/two-step error, but their scales differed sharply. At strength \(.80\), curved median one/two-step NRMSE was \(.0118/.0127\); held-domain conjugacy reached \(.814/1.050\) while decoding remained 1.000. Across both maximum-strength families, 23/30 seed-fold cells crossed at least one closure gate, just below the frozen 24/30 criterion. The label is \texttt{STRUCTURED\_CALIBRATION\_LIMITED}: the test detects a domain-dependent chart, whereas the curved family remains far below the decision region. Later failures cannot be attributed to an implementation that never passes, but one stress family cannot certify universal discriminative power.

\subsection{The h28 transfer gap is action-dominant, not purely entity-specific}\label{the-h28-transfer-gap-is-action-dominant-not-purely-entity-specific}

Frozen h28 state-probe accuracy was \(82.87\%\pm1.81\%\), versus \(20.80\%\pm.20\%\) for the random-label control. Across three datasets, the specific target-minus-wrong intervention effect was \(2.334\pm.231\) logits. Original train-only affine maps had mean error \(.5189\pm.0286\) across 15 action-by-seed cells (cell-level sample SD), with 4/15 below the frozen \(.50\) gate. Translation averaged \(.9166\). Within-test-domain five-fold cross-fit passed every cell at \(.3979\pm.0130\), whereas radial-basis-function ridge averaged \(.5200\) and the source-state-label-gated affine bank \(.5689\). The cross-fit is a descriptive within-domain reference, not a held-entity or deployable map.

Direct tests did not support the stronger entity-specific interpretation. Across seven randomized outer splits, within-entity fitting averaged \(.424\) and cross-entity transfer \(.473\). Per-seed within/cross ratios were \(.895,.900,.894\), above the frozen \(.85\) cutoff, and only 6/21 seed-split directions were favorable. Median same-action/across-entity parameter distances were \(.0281,.0271,.0269\), smaller in every seed than different-action/within-entity distances \(.0292,.0287,.0297\). The frozen label is \texttt{ACTION\_DOMINANT\_GEOMETRY}, not entity conditioning. Action identity dominates this parameter-distance comparison, although entity, token, and contextual variation can still contribute to the transfer gap.

Threshold sensitivity shows where the original result lies. No h28 cell passes at \(.40\), 4/15 pass at \(.50\), and all 15 first pass at \(.58\). The frozen H1 verdict remains \texttt{NOT\ SUPPORTED}. The pattern marks incomplete affine structure near a declared boundary, rather than absence of all action geometry.

\subsection{Early geometry, lexical evidence, and causal depth remain distinct}\label{early-geometry-lexical-evidence-and-causal-depth-remain-distinct}

Mean one-step error was \(.299\) at h4 and \(.347\) at h16; all 15 action-by-seed cells at each layer met the geometric rule. The h28 and h36 means were \(.519\) and \(.549\). Scale alone does not explain this ordering. From h4 onward, mean residual norms were \(11.99\), \(40.23\), \(159.25\), and \(147.73\); displacement norms were \(.686\), \(4.15\), \(36.10\), and \(25.82\). After train-fitted scalar RMS normalization, mean errors remained approximately \(.295,.349,.519,.549\). Effective rank, however, was only \(1.057\) at h4 and \(1.085\) at h16, versus \(1.819\) and \(2.261\) at h28/h36; low-dimensional early geometry remains a live confound.

The h4 carrier remains unresolved by the lexical controls. In the conflict stratum, true-state probe accuracy at h4 was \(.268,.262,.237\), against permuted-label values of \(.192,.219,.184\). This was weakly above control but far below the frozen \(.80\) criterion. Neutral material raised mean h4 action-map error from \(.299\) at distance 0 to \(.392\) at 16 and \(.434\) at 64; no seed passed all five actions at distance 64. The last-16-token static-embedding carrier also missed the all-action rule in every seed. The state-probe and distance gates fail, and the token-only carrier does not pass. The evidence therefore supports neither a lexical explanation nor its exclusion: \texttt{LEXICAL\_ROLE\_UNRESOLVED}.

The causal-depth pattern replicated across three dataset-seed regenerations from the same frozen checkpoint. Mean specific effects were \(.0003\pm.0061\) logits at h4 and \(.0021\pm.0032\) at h16; every within-seed pair-bootstrap interval contained zero. At h28, effects were \(2.424,2.071,2.506\) logits, and at h36 \(4.276,4.399,4.567\). All six intervals excluded zero, and every sign-flip test gave \(p=10^{-5}\). Agreement across all 12 frozen directions yields \texttt{CAUSAL\_DEPTH\_PATTERN\_REPLICATED}. These data establish a replicated sampled-depth dissociation within one checkpoint, not a temporal stage, mediation path, unique carrier, or model-level replication.

Phase 7 separately tested whether the early affine geometry composes. At h4, two-step endpoint error averaged \(.542\) and direct-map discrepancy \(.541\); the h16 values were \(.726\) and \(.717\). No layer-seed cell passed the joint H2 gate (0/6), and held-template endpoint errors exceeded 1.33. Under the lattice, these closure-branch diagnostics remain valid despite the absent causal effect; their label is \texttt{EARLY\_LAYER\_GEOMETRY\_ONLY}, not a causally faithful early operator.

\begin{figure*}[t]
  \centering
  \includegraphics[width=\textwidth]{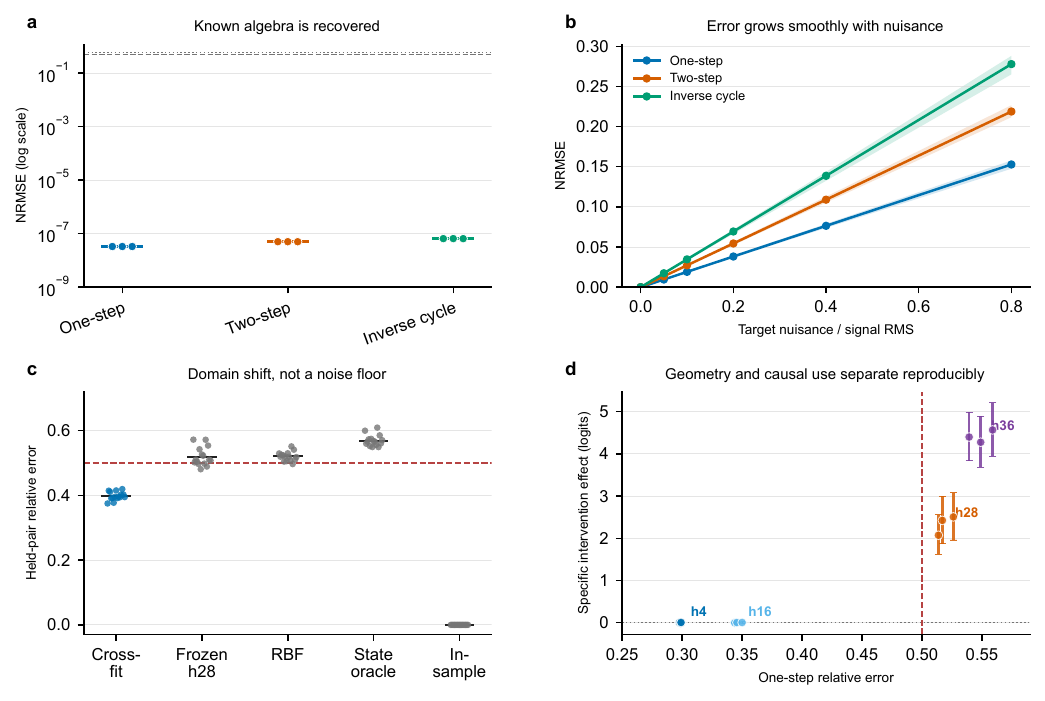}
  \caption{\textbf{Calibration and sampled depth.} \textbf{a}, Exact contextual $S_5$
  carrier results in three held-source folds; lines mark the $.50$ and $.60$ gates.
  \textbf{b}, Mean and empirical 95\% nuisance-seed intervals over 20 seeds after
  averaging folds. \textbf{c}, Fifteen h28 action-by-dataset-seed cells for five
  references; cross-fit is fitted inside the test distribution, and the line is the
  unchanged H1 gate. \textbf{d}, One-step error averaged over five actions within
  each dataset seed versus matched intervention effects and within-seed 95\%
  pair-bootstrap intervals. Each layer has three dataset-seed regenerations of
  160 pairs from the same frozen checkpoint; colors denote layer.}
  \label{fig:reviewer-calibration}
\end{figure*}

\subsection{One-step H1 depends on refitting scope, but composition remains unsupported}\label{one-step-h1-depends-on-refitting-scope-but-composition-remains-unsupported}

Phase 8 initially appeared to show a metric-aligned rescue: weighted full maps refitted on train plus validation averaged \(.4686\), with all 15 cells below \(.50\). There were no exact no-op rows. Displacement stratification still showed denominator sensitivity: unweighted refit error fell from \(.558\) in the smallest-displacement quintile to \(.387\) in the largest; weighted values were \(.533\) and \(.394\).

The outcome-aware descriptive Phase 8b bridge localizes the difference differently. Original selected train maps and full unweighted train maps were numerically identical at \(.5189\) (4/15 below \(.50\)), excluding function-class selection as the explanation. Full weighted train maps reached only \(.5113\) (5/15). Unweighted train-plus-validation refitting, by contrast, reached \(.4744\) (14/15), and weighted refitting reached \(.4686\) (15/15). Refitting improved all 15 cells; median gains were \(.0395\) unweighted and \(.0399\) weighted. Weighting contributed median gains of only \(.0067\) on train and \(.0051\) after refitting, both below the frozen \(.01\) attribution criterion. The label is \texttt{WEIGHTING\_ATTRIBUTION\_NOT\_SUPPORTED}. Within this descriptive bridge, the h28 one-step conclusion is associated mainly with final fitting scope, with a smaller increment from weighting. The older train-only H1 verdict remains frozen.

This sensitivity does not extend to composition. Original h28 composition had mean endpoint error \(.8110\) and direct-map discrepancy \(.7945\). Unweighted refitting changed the pair to \(.7982/.8199\), and weighted refitting to \(.7868/.8059\). Endpoint error improved modestly, but the decisive direct-map gap stayed far above \texttt{.50}; every variant failed H2 in all three datasets. The bridge verdict is \texttt{METRIC\_ALIGNED\_COMPOSITION\_NOT\_SUPPORTED}.

\begin{figure*}[t]
  \centering
  \includegraphics[width=\textwidth]{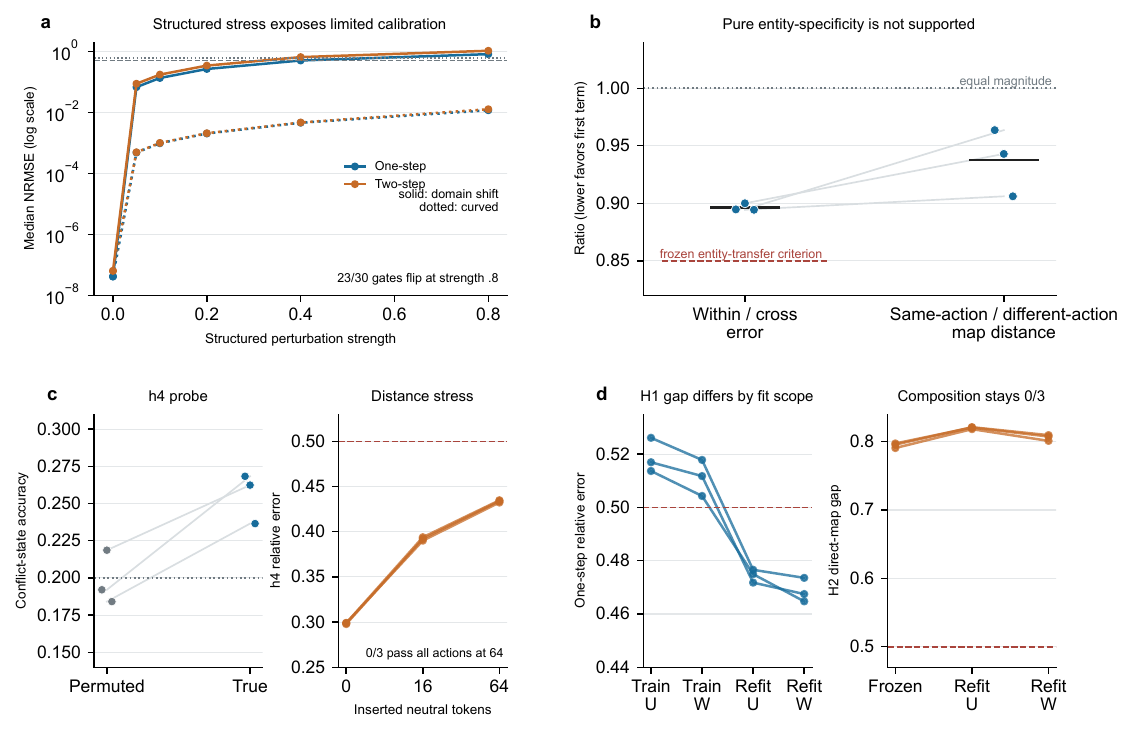}
  \caption{\textbf{Structured diagnostics localize the remaining alternatives.}
  \textbf{a}, Median one- and two-step NRMSE under common curvature (dotted) and
  held-domain conjugacy (solid) across five embedding seeds and three folds.
  \textbf{b}, Three dataset-seed ratios for within/cross-entity error and
  same-action-across-entity/different-action-within-entity map distance.
  \textbf{c}, h4 conflict-state probe versus permuted control and h4 error after
  0, 16, or 64 inserted neutral tokens. \textbf{d}, Seed-level h28 one-step means
  under train/refit and unweighted/weighted fits, alongside the frozen composition
  direct-map gap. U, unweighted; W, weighted. Red lines are frozen gates.}
  \label{fig:reviewer-v2}
\end{figure*}

\subsection{Relative order signals remain weaker than endpoint closure}\label{relative-order-signals-remain-weaker-than-endpoint-closure}

At h28, reversed-order minus correct-order error was \(+.0895\) across ten semantic pairs, with pair-bootstrap 95\% CI \([.0353,.1444]\), one-sided \(p=.00114\), and \(d_z=.960\). Every pair was positive in every dataset seed. Yet correct composition error was \(.811\), a directly fitted two-action map reached \(.443\), and composed-to-direct discrepancy was \(.794\). Applying only the final setter was better than composing both maps (\(.685\)), as expected under last-write absorption. Probe gain was 16.1 percentage points, below 20. H2 remains \texttt{NOT\ SUPPORTED} despite the reproducible order contrast.

The effect was concentrated in particular action families. Empty-versus-fill pairs had an order advantage of \(+.1978\) and a probe advantage of 37.1 points; fill-versus-fill pairs had \(+.0173\) and 2.0 points. H3 showed the same boundary: the same-entity-minus-disjoint commutator difference was \(+.0768\) with 95\% CI \([.0336,.1240]\), but \(d_z=.431\), and only \(44.6\%\) of matched pairs were positive. These decompositions motivate a destructive-clearing versus replacement hypothesis; they do not rescue all-pair H2 or H3.

Grounded H4 remains \texttt{NOT\ TESTED}. Even the best behavior scaffold confirmed the intended zero/one/two-action manipulation only 49.33\% overall, with 28.5\% in its weakest condition, below the 80\% prerequisite. The result constrains that prompt manipulation, not every possible inverse representation.

\subsection{Shared charts and relative laws do not guarantee learned closure}\label{shared-charts-and-relative-laws-do-not-guarantee-learned-closure}

The learned \(S_5\) branch carries more inferential weight than \(\mathbb{Z}_{11}^2\) because all six medium/large trajectories passed task/state eligibility E, endpoint-order D, and shared-chart J. Stable E/D was observed between 6.29M and 12.58M examples, and J at 12.58M for medium and 25.17M for large. These left-truncated observation times do not define a causal emergence sequence.

K was not observed in any trajectory within the fixed \(2^{25}=33.55\) million-example budget, so the event is \texttt{RIGHT-CENSORED}. Final one-step NRMSE was \(1.084\pm.012\) for medium and \(1.002\pm.029\) for large; two-step values were \(1.070\pm.010\) and \(1.022\pm.027\). Nested frozen-weight diagnostics found seen-source one-step NRMSE \(.260\pm.006\) versus held-source \(.786\pm.036\), held composition \(.835\pm.022\), teacher-forced composition \(.540\pm.015\), and a separately fitted depth-one map \(.497\pm.030\). Source-state extrapolation and closed-loop accumulation both matter, but neither rank 119 nor residual multilayer perceptrons consistently rescue the result.

We retain the \(\mathbb{Z}_{11}^2\) branch as a task-ineligible diagnostic. Medium/large ID accuracies exceed 99\%, but length-OOD accuracy is only 6.5\%--7.3\%, and loop/route scores remain near chance. All nine final models nevertheless achieve commuting-pair AUROC 1.000 while the absolute H1, H2, and H4 metrics fail. Relative pair discrimination can coexist with poor endpoints, but this branch cannot adjudicate closure in an algorithmically qualified model.

The preregistered H5 composite also failed as a measurement. Across 99 nonzero checkpoints, partial Spearman \(\rho=.0911\) with trajectory-cluster 95\% CI \([-.0352,.2486]\) did not support the predicted association. One trajectory illustrates why: at initialization, it decoded states at 3.34\% while H1/H2 errors were spuriously low at \(.091/.120\); after state differentiation, decoding reached 89.50\% and the errors rose to \(.813/.920\). Representation scale and an unmet task gate confounded the mixed composite. No favorable subset or alternative composite replaces it.

\begin{table*}[t]
  \centering
  \caption{\textbf{Principal frozen and follow-up verdicts.} Diagnostic
  labels remain separate from earlier frozen decisions.}
  \label{tab:verdicts}
  \footnotesize
  \setlength{\tabcolsep}{3.5pt}
  \renewcommand{\arraystretch}{1.12}
  \begin{tabularx}{\textwidth}{@{}>{\raggedright\arraybackslash}p{0.15\textwidth}>{\raggedright\arraybackslash}p{0.20\textwidth}>{\raggedright\arraybackslash}p{0.21\textwidth}X@{}}
    \toprule
    Setting & Test & Verdict & Decisive evidence \\
    \midrule
    Exact carrier & Zero/support recovery & POSITIVE CONTROL RECOVERED
      & All formal/reproduction/support folds pass near numerical precision. \\
    Exact carrier & Structured stress & STRUCTURED CALIBRATION LIMITED
      & Monotone errors, but 23/30 rather than 24/30 gate flips. \\
    Qwen3-4B h28 & Frozen H1 train-only & NOT SUPPORTED
      & Mean $.519$; 4/15 below $.50$; cross-fit is descriptive. \\
    Qwen3-4B h28 & Entity/context diagnostic & ACTION-DOMINANT GEOMETRY
      & Within/cross ratio about $.90$; action distance dominates. \\
    Qwen3-4B h4 & Lexical diagnostic & LEXICAL ROLE UNRESOLVED
      & Weak conflict probe; distance and token-only controls do not adjudicate. \\
    Sampled layers & Paired interventions & CAUSAL DEPTH PATTERN REPLICATED
      & h4/h16 intervals include zero; h28/h36 positive in all datasets. \\
    Qwen3-4B h28 & H1 attribution & WEIGHTING ATTRIBUTION NOT SUPPORTED
      & Refit gain about $.040$; weighting gain only $.005$--$.007$. \\
    Qwen3-4B h28 & Refit-map H2 bridge & NOT SUPPORTED
      & All variants 0/3; direct-map gaps $.794$--$.820$. \\
    Qwen3-4B h4/h16 & Layer-local H2 & EARLY LAYER GEOMETRY ONLY
      & One-step geometry passes, but frozen H2 is 0/6. \\
    Learned $S_5$ & K held-source closure & RIGHT-CENSORED
      & J passes 6/6; K absent through the fixed 33.55M-example budget. \\
    $\mathbb{Z}_{11}^2$ trajectories & H5 algebra-to-coherence & NOT SUPPORTED / NOT INTERPRETABLE
      & CI crosses zero; task gate fails; early collapse invalidates low errors. \\
    \bottomrule
  \end{tabularx}
\end{table*}

\section{Discussion and Limitations}\label{discussion-and-limitations}

Under the evidence lattice, the layer results are different partial outcomes rather than contradictions. h4/h16 are geometry-positive and causal-use-negative; h28 is causal-use-positive and frozen-closure-negative. Neither region licenses the conjunction. Selecting one layer because it decodes well or patches strongly would instead turn a local result into a model-wide mechanism.

The positive carrier and structured stresses answer different calibration questions. Near-zero recovery validates the fit, chart, holdout, and gates. Target nuisance provides a graded specificity check, whereas held-domain conjugacy moves the gates into the grounded error range through structured chart mismatch. Because the curved family remains far below those gates, the stress suite does not establish universal statistical power. That limitation is part of the result.

The h28 transfer gap does not warrant a pure entity-binding account. Within-domain cross-fit shows attainable local fit but has access to the evaluation domain. Randomized splits yield only modest within-entity gains, and the frozen map comparison is action-dominant. The supported description is a distribution-dependent map with action-structured parameters; entity, token, template, and contextual effects remain entangled.

Specification matters for the one-step result. The observed h28 H1 estimate changes far more between train-only and train-plus-validation fits than between weighting choices, though neither procedure uses test data. The refit is a legitimate generalization estimate under a different protocol, not a post hoc revision of the original decision. Its failure to carry over to H2 sets the boundary: better one-step interpolation does not imply reusable composition.

We distinguish confirmatory from descriptive analyses. The prospectively frozen set comprises the original H1--H5 gates, the Phase 6 positive carrier and target-nuisance tests, Phase 7 early-layer H2, and the Phase 8 structured, entity, lexical, metric, geometry, and causal-replication families. Phase 8b is outcome-aware and descriptive; semantic-family decompositions, frozen-weight mechanism grids, and sampled-layer geometry controls remain hypothesis-generating. Depending on the branch, the independent unit is the dataset seed, complete training trajectory, or held-source fold specified in Methods. Prompts, actions, checkpoints, and grid cells are not independent model replicates.

Table~\ref{tab:carrier-coverage} makes the limited carrier coverage explicit.

\begin{table*}[t]
  \centering
  \caption{\textbf{Carrier coverage.} Pass and not supported refer only to the
  listed test and frozen representation.}
  \label{tab:carrier-coverage}
  \footnotesize
  \setlength{\tabcolsep}{4pt}
  \renewcommand{\arraystretch}{1.12}
  \begin{tabularx}{\textwidth}{@{}>{\raggedright\arraybackslash}p{0.24\textwidth}>{\raggedright\arraybackslash}p{0.13\textwidth}>{\raggedright\arraybackslash}p{0.15\textwidth}>{\raggedright\arraybackslash}p{0.13\textwidth}X@{}}
    \toprule
    Candidate carrier & One-step & Composition & Causal use & Current status \\
    \midrule
    Final-token residual h4 & pass & not supported & not detected &
    Low-dimensional geometry; lexical role unresolved \\
    Final-token residual h16 & pass & not supported & not detected &
    Layer-local affine geometry only \\
    Final-token residual h28 & fit-scope sensitive & not supported & replicated positive &
    Causally usable signal without global affine closure \\
    Final-token residual h36 & not supported & not tested & replicated positive &
    Causal-use branch only \\
    Last-16 static token embeddings & not supported & not tested & not tested &
    Does not explain h4 by itself \\
    Other tokens/cross-layer spans & not tested & not tested & not tested & Live alternative \\
    Attention/MLP paths & not tested & not tested & not tested &
    Requires path-specific natural targets \\
    KV cache/distributed state & not tested & not tested & not tested &
    Requires a different carrier and metric \\
    \bottomrule
  \end{tabularx}
\end{table*}

The grounded evidence is restricted to one post-trained model family, an absorbing setter domain, and four sampled final-token layers. The conflict probe is weak, and entity and context factors are correlated. Causal replication uses three generated datasets from the same checkpoint, not independently pretrained checkpoints. In the learned world, K is right-censored. None of the results excludes nonlinear, attention-mediated, cross-token, cross-layer, KV-cache, or context-conditioned operators.

Even with those limits, the measurement program suggests a reporting standard. Establish state availability, then test causal use and held-source closure as separate branches. Report natural endpoints and continuous error, with calibration against a known operator under structured misspecification. Keep within-domain reference fits distinct from held-domain mechanisms. Composition tests should include direct maps, last-action controls, commutativity, and inverse cycles. Behavioral eligibility, sufficient statistics, manifests, and independent audits complete the record.

\section{Conclusion}\label{conclusion}

The held-source measurement recovers a known affine action algebra, while structured domain mismatch can move its gates into the empirical failure range. In Qwen3-4B, action-conditioned geometry, state decoding, and within-checkpoint dataset-seed causal-use results appear at different sampled depths. Direct entity tests do not support a purely entity-specific account; lexical controls remain unresolved, and the observed h28 one-step result varies substantially with final fitting scope. The boundary is composition: neither early-layer nor h28 refit maps satisfy the unchanged composition construct. Learned finite worlds draw the same distinction between relative structure or shared charts and absolute held-source closure.

The positive claim is deliberately bounded. Within the tested carriers, state availability, causal local use, action-structured geometry, and reusable affine closure are distinct empirical constructs. Calling an internal map a causally faithful reusable operator should require their conjunction in the same eligible carrier. Richer internal operators remain possible; this study specifies the evidence such a claim must reconstruct.

\section*{Data Availability}

Source rows for Figures 2, 3, S1--S4 accompany this arXiv version under
\texttt{anc/source\_data/}, with result, manifest, audit, and source-table SHA256
values in the associated metadata JSON files. A compact reproducibility snapshot
is provided as \texttt{anc/reproducibility\_bundle.zip}. No human-participant or
personal data were collected. Large model weights, raw activation tensors,
intermediate checkpoints, and multi-gigabyte fitted artifacts are not redistributed.

\section*{Code Availability}

The ancillary reproducibility bundle contains preregistrations, frozen
configurations, experiment code, remote wrappers, independent auditors, plotting
scripts, tests, environment specifications, audit reports, and source tables.
The third-party \texttt{state-probes} submodule is identified by its upstream URL
and pinned commit but is not redistributed. Excluded large artifacts can be
regenerated from the recorded model identifiers and frozen configurations; their
hashes and passports remain in the included reports and metadata.

\section*{Ethics Declaration}

This study used pretrained and from-scratch computational models, procedurally generated prompts, and finite synthetic transition systems. It involved no human participants, personal data, clinical data, or animal research.

\section*{Author Contributions}

Dekun Yang: Conceptualization, Methodology, Software, Validation, Formal analysis, Investigation, Data curation, Visualization, Writing -- original draft, Writing -- review and editing.

\section*{Competing Interests}

The author declares no competing interests.

\section*{Funding}

No external funding was received for this work.

\section*{AI-Assistance Disclosure}

OpenAI Codex assisted with code generation, experiment orchestration, validation scripts, figure production, evidence organization, and language drafting. Experimental claims and numerical values were checked against versioned machine-readable artifacts and independent audit outputs. The authors remain responsible for the scientific design, interpretation, and final text.

\setcounter{figure}{0}
\renewcommand{\thefigure}{S\arabic{figure}}
\begin{figure*}[t]
  \noindent{\large\bfseries 8. Supplementary Results}\par
  \vspace{0.35em}
  \noindent{\normalsize\bfseries 8.1 Threshold sensitivity without verdict
  movement}\par
  \vspace{0.45em}
  \centering
  \includegraphics[width=\textwidth]{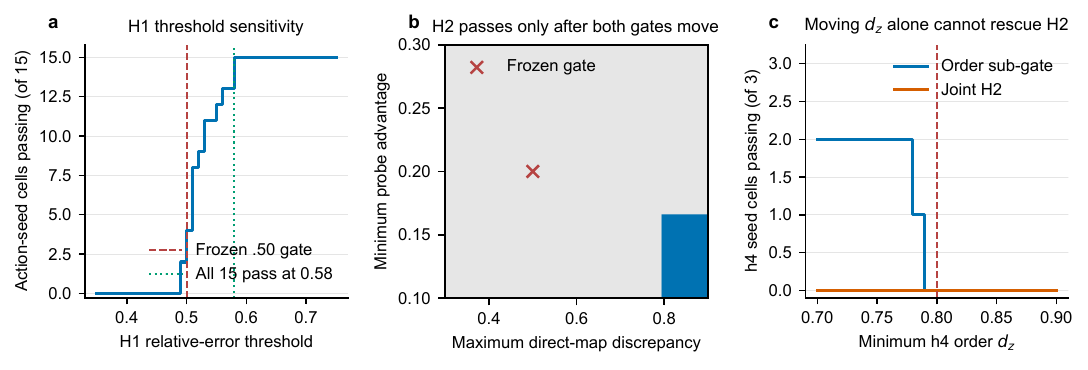}
  \caption{\textbf{Frozen thresholds expose different degrees of boundary
  sensitivity.} \textbf{a}, Number of 15 h28 action-by-seed cells below each H1
  threshold. The red line is the frozen $.50$ gate; all cells first pass at $.58$.
  \textbf{b}, H2 verdict over direct-map-discrepancy and probe-gain thresholds while
  other gates stay fixed; the cross is the frozen $(.50,.20)$ point. \textbf{c},
  number of Phase 7 h4 seed cells passing the order sub-gate or full H2 as the
  minimum $d_z$ varies. Relaxing $d_z$ alone never changes joint H2 because direct
  and probe gates still fail. No panel changes an original decision.}
  \label{fig:threshold-sensitivity}
\end{figure*}

\begin{figure*}[t]
  \noindent{\normalsize\bfseries 8.2 Failed H5 composite and collapse
  diagnostic}\par
  \vspace{0.45em}
  \centering
  \includegraphics[width=\textwidth]{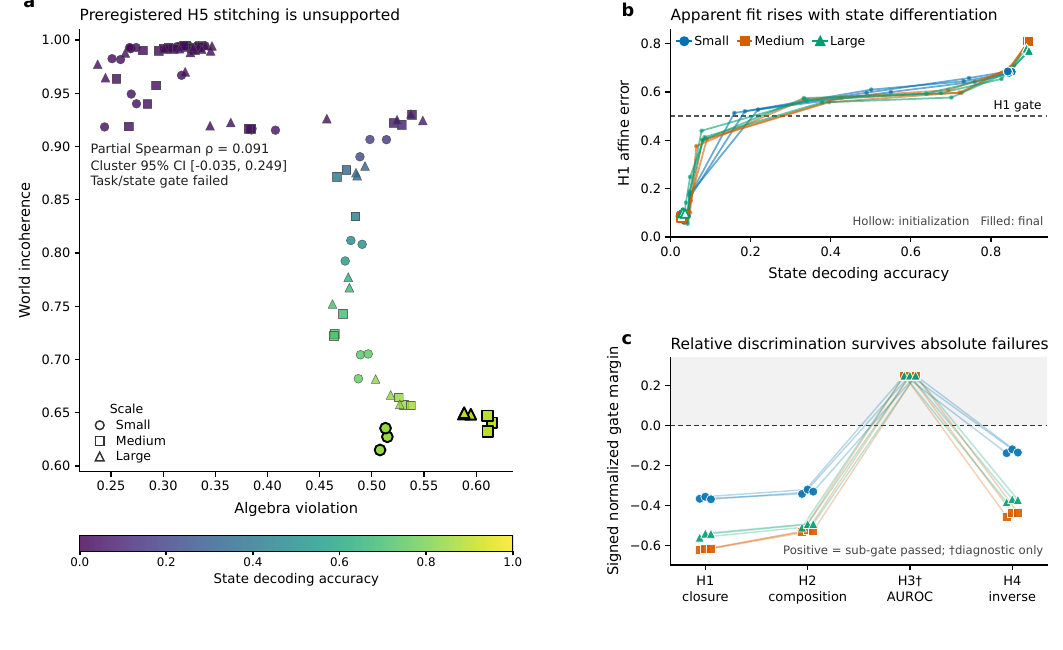}
  \caption{\textbf{State differentiation exposes failure of the preregistered H5
  composite.} \textbf{a}, Algebra violation and world incoherence across 99 nonzero
  checkpoints; partial Spearman $\rho=.091$ with trajectory-cluster 95\% CI
  $[-.035,.249]$, and the task gate fails. \textbf{b}, One trajectory shows H1
  error rising as collapsed state representations differentiate. \textbf{c}, Signed
  margins use $(.5-\mathrm{H1})/.5$, $(.6-\mathrm{H2})/.6$,
  $(\mathrm{AUROC}-.8)/.8$, and $(.5-\mathrm{H4})/.5$. H3 remains diagnostic
  and cannot override absolute gates.}
  \label{fig:h5-stitching}
\end{figure*}

\begin{figure*}[t]
  \noindent{\normalsize\bfseries 8.3 Layer-local composition gates}\par
  \vspace{0.45em}
  \centering
  \includegraphics[width=\textwidth]{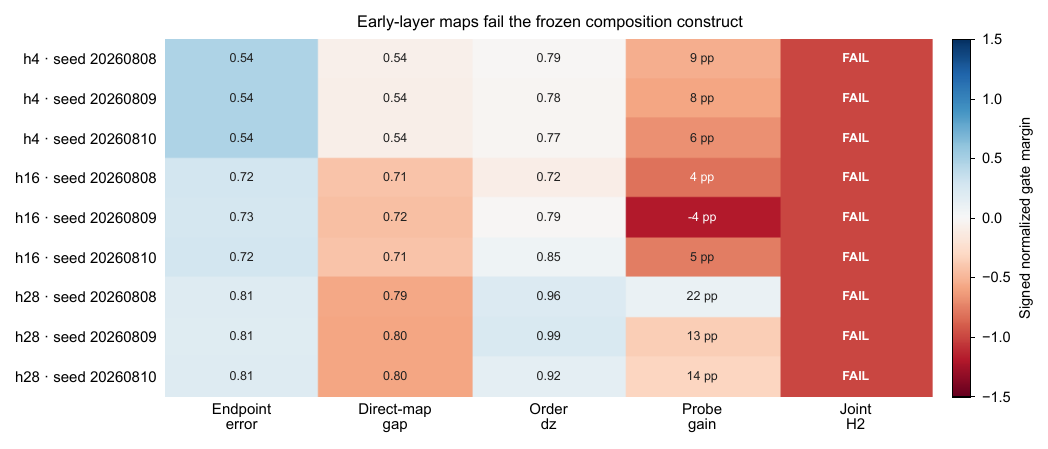}
  \caption{\textbf{Early-layer one-step geometry does not satisfy the frozen
  composition construct.} Raw cell text reports endpoint error, direct-map gap,
  order $d_z$, probe gain, and joint H2 status. Color is signed normalized distance
  from the frozen sub-gate. h28 rows are frozen compatibility comparators. No
  layer-seed cell passes joint H2.}
  \label{fig:layer-composition}
\end{figure*}

\begin{figure*}[t]
  \noindent{\normalsize\bfseries 8.4 Layer scale and effective dimension}\par
  \vspace{0.45em}
  \centering
  \includegraphics[width=\textwidth]{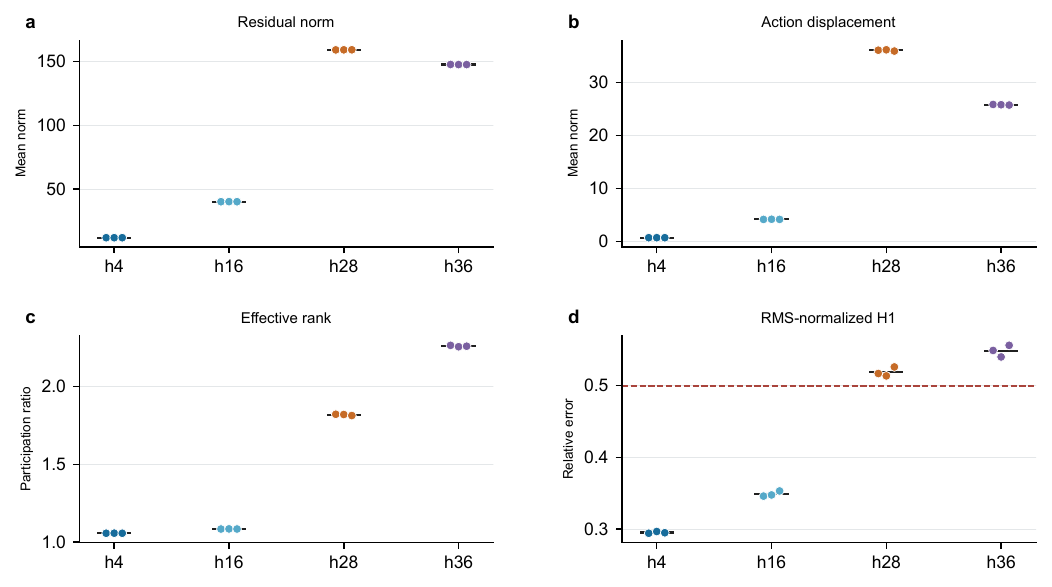}
  \caption{\textbf{Scalar scale control preserves the sampled-depth ordering,
  while early representations are low dimensional.} \textbf{a}, Mean residual
  norm. \textbf{b}, Mean action displacement norm. \textbf{c}, Covariance
  participation ratio. \textbf{d}, One-step error after train-fitted scalar RMS
  normalization. Points are three activation datasets; black bars are seed means.}
  \label{fig:layer-geometry}
\end{figure*}

\FloatBarrier
\begingroup
\footnotesize
\setlength{\bibsep}{0pt plus 0.15ex}
\bibliographystyle{plainnat}
\bibliography{references}
\endgroup

\end{document}